# Boundary identification of events in clinical named entity recognition

*Azad Dehghan*
School of Computer Science
The University of Manchester
Manchester, UK
a.dehghan@cs.man.ac.uk

*Abstract*—The problem of named entity recognition in the medical/clinical domain has gained increasing attention due to its vital role in a wide range of clinical decision support applications. The identification of complete and correct term span is critical for further knowledge synthesis (e.g., coding/mapping concepts thesauruses and classification standards). This paper investigates boundary adjustment by sequence labeling representations models and post-processing techniques in clinical named entity recognition (recognition of clinical events). Using current state-of-the-art sequence labeling algorithm (conditional random fields), we show experimentally that sequence labeling representation and post-processing can be significantly helpful in determining exact boundaries of clinical events.

*Keywords—clinical named entity recognition; boundary identification; clinical concept extraction*

## I. INTRODUCTION

The problem of named entity recognition and classification (NERC or NER) is concerned with identification and (coarse-grained) classification of proper names or names of things, either physical or conceptual. In the medical or clinical domain, NER typically constitutes names of concepts e.g., *Problem* (sub-classes include e.g., disease or syndrome, anatomic abnormality, sign or symptom, etc.), *Treatment* (e.g., therapeutic or preventive procedure, clinical drug, medical device, etc.), *Test* (e.g., laboratory/diagnostic procedures, etc.), and so forth.

NER research has attracted increasing attention due to its vital role in many real-world applications such as clinical decision support systems (CDSS). Since 2010, a research driver in this domain has been the NIH-funded Informatics for Integrating Biology and the Bedside (i2b2) who has organized three NER-related tasks [1, 2, 3, 4]. i2b2 has also provided high-quality corpora related to multiple areas of clinical domain, including NER, co-reference resolution, sentiment analysis, temporal expressions recognition and normalization, and temporal relations extraction. In addition, the Conference and Labs of the Evaluation Forum (CLEF) recently organized an NERC task (CLEF-ER 2013) focused on NER and fine grained classification (or mapping of concepts to medical thesaurus: UMLS/SNOMED-CT) of medical concept in (multi-lingual) biomedical corpora.

Challenges in clinical NERC have mainly been identified as [5]: term ambiguity/complexity, abbreviation and acronym resolution, exact boundary identification, fine-grained classification, and data availability and quality.

This paper investigates exact boundary identification of events in the clinical domain—by exploring sequence model representation and post-processing boundary adjustment.

The reminder of this paper is organized as follows: *Section II* refers the reader to recent work in clinical NER, *Section III* describes experimental design, *Section IV* gives an overview of experimental results, *Section V* provides discussion of obtained results, *Section VI* summarizes and concludes results and discussions and lastly, Section VII is the appendices.

## II. RELATED WORK

For background, in particular recent work in clinical NER, the interested reader is referred to [1, 2, 3, 4, 6, 7].

## III. EXPERIMENTAL DESIGN

This section describes experimental design in the context of the study objective described in Section I. Experiments conducted and described in this paper are motivated by investigating: (1) boundary identification through token sequence representation modeling (e.g., IO, IOB, IOBW, IOBEW.) and (2) boundary adjustment by post-processing.

Sequence model representations used for NER in the biomedical domain is largely dominated by the BIO and less so IO models (labels; I—refers to *inside*, O—*outside*, B—b*eginning*, (E—*end*, W—*single word/token term*)). However, the motive for such approaches appears to be omitted in literature. Hence, this provides a motivation to investigate sequence model representations and its application in boundary adjustment.

### A. Initial hypotheses

A set of initial hypothesis were formulated in context of described motivation:

1) There is no difference in terms of discriminative power between IO, IOB, IOBW, IOBEW sequence labeling models, in particular, with regards to exact boundary identification.

2) Boundary adjustment post-processing has no impact in strict boundary/span identification of clinical events.

*B. Design*

All experimental results were obtained using 5 x 5-fold cross validation, with folds across events being unpaired. Java (class.method) Collections.shuffle() was used as a source of randomness.

A balanced designed ANOVA analysis was used to detect statistical significant variation across models. Subsequently, two-tailed unpaired T-test was applied to confirm statistical variation among individual models. Given assumptions (ANOVA and T-test) of the data were also analyzed (e.g., normality, variance differences, etc.) to confirm suitability and reliability of statistical analysis. Statistical analysis was carried out using GNU S or R.

*C. System*

The Clinical NERC Toolbox[1] (which uses the General Architecture for Text Engineering (GATE) framework[2] [10]), was used in all experiments.

The implementation of conditional random fields (CRF) used was CRF++ 0.58 (Java SWIG interface)[3]; with the following parameters: C=1.00, ETA: 0.0001 and L2-regularization algorithm.

The JAPE grammar was used for rule engineering[4].

*D. Architecture and methods*

FIGURE I. SYSTEM PIPELINE

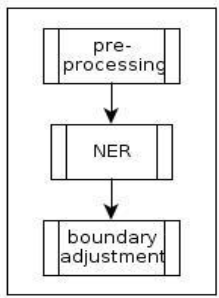

The system pipeline is made up of the following components; *nota bene*: processing components (represented as bullet points) are given in correct sequential ordering (Fig. 1):

*1) Pre-processing:*
- ANNIE Tokenizer
- ANNIE Sentence splitter
- OpenNLP Part-of-speech (POS) tagger
- OpenNLP Chunker
- Porter's stemmer

*2) NER:*
- ML prediction module(s) (IO, IOB, IOBW)
- Label fixer (see example in Table I)

TABLE I. LABEL FIXER HEURISTIC

| # | Initial predictions | prediction correction |
|---|---|---|
| a | ... O O O I I I I ... | ... O O B I I I I ... |
| b | ... O O O B O O O ... | ... O O O B I O O ... |

*3) Boundary adjustment:*
- *Boundary label adjustment* - see Table II.
- *Boundary expander* - by including adjacent tokens to the right and left of events that possess POS/chunk tags (that corresponded to nouns and noun phrases and their constituents), and determiners (e.g., "a", "this", "her", etc.).

TABLE II. BOUNDARY LABEL ADJUSTMENT

| # | Initial predictions | prediction correction |
|---|---|---|
| c | ... O O O B O I I ... | ... O O O B I I I ... |
| d | ... O O O B I I B I I ... | ... O O O B I I I I I ... |
| e | ... O O O B I I B I I B I ... | ... O O O B I I I I I I I ... |

*E. Features*

While feature bias may be relevant when considering domain peculiarities, the features set finally selected for building the explored models have repeatedly shown to provide best discriminative power [1, 2].

A combination of the forward and backward feature selection approaches was adopted and a total of 31 features were selected. These features may be clustered into three sets with 7 feature groups across (see following list). The feature space is made up of (for each feature group x, x: 1-7): $(fg_x)$, $(fg_{x+1})$, $(fg_{x+2})$, $(fg_{x-1})$, $(fg_{x-2})$ and an additional n-gram feature consisting of specific feature-groups $(fg_1+fg_2+fg_5)$.

*1) Textual*
- $fg_1$: Token: alpha numeric character sequence.
- $fg_2$: Stem of Token: the stem of each token (using Porter's stemmer)[5].

*2) Syntactic*
- $fg_3$: POS: the POS-tags for each token (using OpenNLP[6] with cTAKES model[7]).
- $fg_4$: Shallow parse: the chunk tag of each token (using OpenNLP with cTAKES model).

*3) Orthographic*

---
[1] Clinical NERC: http://clinical-nerc.sourceforge.net/
[2] http://www.gate.ac.uk
[3] http://code.google.com/p/crfpp/
[4] http://gate.ac.uk/sale/thakker-jape-tutorial/
[5] http://tartarus.org/martin/PorterStemmer/
[6] http://opennlp.apache.org/
[7] http://ctakes.apache.org/

- $fg_5$: Token kind: i.e., kind corresponds to {word, number, symbol, punctuation}
- $fg_6$: Token-case: i.e., {lower case, upper case, upper initial, mixed caps, all caps}

*4) Feature group n-gram*
- $fg_7$: features n-gram: i.e., $fg_1$, $fg_2$, and $fg_5$ concatenated to create feature n-grams.

The CRF++ template used follows:

| #String | #Stem | #POS |
|---|---|---|
| U00:%x[-2,1] | U05:%x[-2,2] | U10:%x[-2,3] |
| U01:%x[-1,1] | U06:%x[-1,2] | U11:%x[-1,3] |
| U02:%x[0,1] | U07:%x[0,2] | U12:%x[0,3] |
| U03:%x[1,1] | U08:%x[1,2] | U13:%x[1,3] |
| U04:%x[2,1] | U09:%x[2,2] | U14:%x[2,3] |

| #Chunk | # Ortho:TokenKind | #Ortho:TokenCase |
|---|---|---|
| U15:%x[-2,4] | U20:%x[-2,5] | U25:%x[-2,6] |
| U16:%x[-1,4] | U21:%x[-1,5] | U26:%x[-1,6] |
| U17:%x[0,4] | U22:%x[0,5] | U27:%x[0,6] |
| U18:%x[1,4] | U23:%x[1,5] | U28:%x[1,6] |
| U19:%x[2,4] | U24:%x[2,5] | U29:%x[2,6] |

**#String/Stem/TokenKind**
U30:%x[0,1]/%x[0,2]/%x[0,5]

*F. Models*

Four models or configurations were investigated:

1) IO
2) IOB
3) IOBW
4) IOBW+ (IOBW and boundary adjustment post-processing)

*G. Corpus Profile*

The primary corpus used for the experimentation is the i2b2 2012 events corpus (combined training and test set: 310 documents). Further validation of our methods is derived from a combined dataset of the i2b2 2010 and 2012 event corpora (data available here[8]). In total, 616 documents are used for training and 120 documents as an unseen test set (results not included; see Clinical NERC).

In order to profile the primary corpus an analysis of events that includes word-distribution, lexical variation (proportion of unique words) and acronyms/abbreviation distribution was completed (see detailed results in Section VII: Table I and Table II).

The word-distribution was calculated by normalizing concepts (using UMLS LVG) by removing determiners,

---

[8] https://www.i2b2.org/NLP/DataSets/Main.php

punctuation, possessives; terms were lowercased and alphabetically sorted. Acronym and abbreviation distribution was semi-automatically counted by employing a set of heuristics and gazetteers.

FIGURE I. WORD DISTRIBUTION

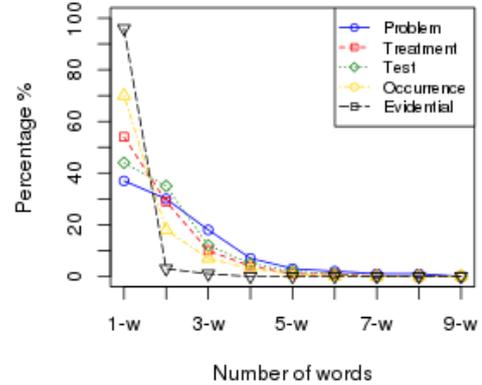

Evidential events were discovered as outliers in this analysis: 96% of the terms are made up of a single word; in addition, occurrence events followed suit with 70% of terms made up of a single word (see Fig. I). Further, evidential events were made up of only 10% of unique words (occurrence: 30%). These two categories stood also out with regards to proportion of acronyms and abbreviations (APPENDIX: Table II).

### IV. RESULTS

The probable bias derived from experimental design including dataset (e.g., only clinical data type used: discharge summary), features set, and domain peculiarities (e.g., linguistic profile of events) ought to be taken into consideration when interpreting findings. However, methodological findings show that methods investigated can to a large extent be generalizable within the clinical or medical domain.

*A. Boundary adjustment: sequence model selection*

Initial investigation into various sequence labeling models such as IO, IOB, and IOBW showed obvious prediction errors. Consequently, a simple heuristic (Table I: *Label fixer*) was developed to correct these. Reference [12] partially applied this method (Table I: #a) to a temporal expressions identification task as part of post-processing. However, *Label fixer* is not considered to be part of boundary adjustment post-processing in this study.

Experimental data showed that IOB and IOBW models performed (in terms of strict scores) consistently and significantly better than IO models. However, the exception was the evidential category which showed no significant variation across models (see Fig. II). In addition, no significant difference was discovered between IOB and IOBW models across all events (when considering exact evaluation criteria). However, slight, but non-significant, higher scores were observed for IOBW models when using larger datasets.

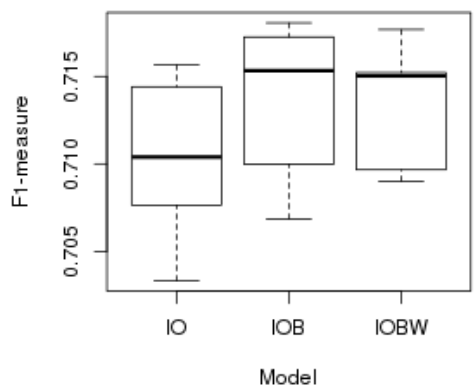

FIGURE II. EVIDENTIAL MODELS

For problem, occurrence, treatment and test events, IOB/IOBW models showed statistically significant increase compared to baseline IO models (only IO vs. IOBW T-test results are shown; Table III). An increase of approximately 4% improvement in strict F1-measure compared to corresponding IO models was observed (see averaged 5x5-fold F1-measure in Table IV). Consequently, the IOBW label sequence representation was preferred for boundary identification.

TABLE III. T-TEST RESULTS

| Event | Models |
|---|---|
| | IO vs. IOBW |
| Problem | $8.6 \times 10^{-9}$ |
| Test | $1.8 \times 10^{-6}$ |
| Treatment | $1.2 \times 10^{-9}$ |
| Occurrence | $6.2 \times 10^{-5}$ |

TABLE IV. ANOVA RESULTS

| Models (averaged exact F1-measure) | | | | | |
|---|---|---|---|---|---|
| Event | IO | IOB | IOBW | IOBW+ | p-value |
| Problem | 0.5803 | 0.6192 | 0.6183 | 0.6658 | $4.18 \times 10^{-11}$ |
| Test | 0.6340 | 0.6813 | 0.6836 | 0.7277 | $2 \times 10^{-16}$ |
| Treatment | 0.6205 | 0.6662 | 0.6687 | 0.7035 | $2 \times 10^{-16}$ |
| Occurrence | 0.564 | 0.580 | 0.580 | N/A | $1.8 \times 10^{-6}$ |
| Evidential | 0.710 | 0.713 | 0.713 | N/A | 0.485 |

B. *Boundary adjustment: post-processing*

Three event categories were selected to be included in the boundary-adjustment post processing investigation (i.e., problem, treatment and test).

The boundary adjustment post-processing module was developed using a rule-based approach. This module accounts for:

(i.) *Boundary label adjustment* (see Table II)

(ii.) *Boundary expander* (see Section III:D)

IOBW+ showed a significant increase compared to corresponding baseline IOBW models. IOBW+ exhibited an increase of approximately 5% increase compared to corresponding IOBW models and a 9% increase over IO models. These differences were confirmed as strongly significant (Table V).

TABLE V. T-TEST RESULTS

| Event | Models |
|---|---|
| | IOBW vs. IOBW+ |
| Problem | $8 \times 10^{-8}$ |
| Test | $2.1 \times 10^{-10}$ |
| Treatment | $9.4 \times 10^{-9}$ |

C. *Lenient results*

The aim of this study was primarily to inform as to the best approach for exact boundary identification for NERC. However, lenient scores show interesting characteristics and were therefore further investigated.

Significant variation in lenient evaluation scores between IO, IOB, IOBW, and IOBW+ model representations were observed using a balanced ANOVA (Table VI). In addition, IO models seem to be the slightly better approach when using lenient matching.

TABLE VI. ANOVA RESULTS

| Models (averaged lenient F1-measure) | | | | | |
|---|---|---|---|---|---|
| Event | IO | IOB | IOBW | IOBW+ | p-value |
| Problem | 0.8192 | 0.8125 | 0.8113 | 0.8266 | $2.09 \times 10^{-10}$ |
| Test | 0.8326 | 0.8226 | 0.8239 | 0.8312 | $3.71 \times 10^{-8}$ |
| Treatment | 0.8107 | 0.8021 | 0.8058 | 0.81 | $1.26 \times 10^{-7}$ |
| Occurrence | 0.6834 | 0.6811 | 0.6767 | N/A | 0.0079 |
| Evidential | 0.7158 | 0.7159 | 0.7161 | N/A | 0.99 |

Furthermore, pair-wise t-tests were conducted among models to confirm statistical significance of variation of inexact or lenient results (see Table VII.).

TABLE VII. PAIR-WISE T-TEST

| Pair-wise T-test results (p-value) | | | |
|---|---|---|---|
| Event | IO vs. IOB | IO vs. IOBW | IOB vs. IOBW |
| Problem | $7.9 \times 10^{-5}$ | $3.8 \times 10^{-4}$ | 0.3633 |
| Test | $1.1 \times 10^{-4}$ | $9.5 \times 10^{-5}$ | 0.07602 |
| Treatment | $6.3 \times 10^{-5}$ | $3.1 \times 10^{-4}$ | 0.3037 |
| Occurrence | 0.2375 | 0.009 | 0.02429 |
| Evidential | 0.9718 | 0.8884 | 0.923 |

As confirmed by multiple pair-wise t-tests carried out, IO sequence model representation shows statistically significantly better scores than IOB and IOBW models when evaluated with lenient criteria (i.e., for event categories: problem, test and treatment). This is in clear contradiction to exact criteria where IO models demonstrate significantly lower scores to IOB and IOBW models.

Further, no statistical variation was discovered among IOB and IOBW models, except for the occurrence category.

## V. DISCUSSION

Experimental results obtained have showed both anticipated and unexpected findings. Firstly, considering our initial hypotheses (refer to Section III):

(1) IO/IOB/IOBW sequence labeling models showed statistically significant variation both when considering strict/exact and lenient/inexact evaluation criteria. Hence, the null hypothesis was rejected.

(2) Boundary adjustment post-processing showed significant gains compared to the baseline models. Hence, the null hypothesis was again rejected. This finding in particular was expected as results from previous work [13] were indicative of the results obtained.

Moreover, reflecting on results obtained with regards to our first hypothesis, IO/IOB/IOBW sequence representation models showed significant variation (across 4 out of the 5 events considered, the evidential category was the outlier) when considering both exact and inexact evaluation criteria. When considering exact span identification of a clinical event, our experimental results indicate that a preference of IOB and IOBW (as opposed to IO) model representation should be adopted. However, when considering lenient evaluation criteria, IO models showed contradictory behavior. IO models showed slight but statistically significantly better scores when considering inexact criteria. This may indicate that IO model representation combined with boundary adjustment may be the best approach even for strict boundary identification.

Further, no variation was discovered between evidential category sequence model representations, i.e., IO, IOB and IOBW (see Fig. II and Table IV). Analysis of the corpus/events shows that evidential events are largely made up of single word terms with extremely low lexical diversity; this may be why different model representations showed no significance difference in this case. Evidential events were also the easiest to predict (that may also be explained by its characteristics: i.e., lexical diversity and word distribution).

Moreover, the IOBW sequence model representation was initially hypothesized to provide better discriminative power (than IOB) due to its hypothesized ability to differentiate between multi-word terms and single word term. However, this premise proved false: no significant difference was found between IOB and IOBW models when considering strict evaluation criteria (and only one event exhibited statistical significant variation when considering lenient criteria i.e., occurrence).

## VI. CONCLUSION

This paper has investigated and analyzed exact boundary identification of clinical events.

The experiments conducted in this study support current premises or assumptions (e.g., bias towards IO and IOB model representation in the biomedical domain) or appear to be novel findings worthy of further work (sequence model representation preference depending on aim i.e., exact vs. lenient boundary identification and significance of boundary adjustment to exact span identification):

- IOB and IOBW model representation provide better discriminating ability (than IO) to determine the exact span boundary. (Events considered: problem, treatment, test, occurrence, and evidential).

- IO sequence model representation showed slight but significantly better performance when evaluated with lenient measures. (Events considered: problem, treatment, and test).

- IOB vs. IOBW models showed no significant variation across most event types investigated (considering both lenient and exact measures). The exception was the occurrence category when considering lenient measure. (Events considered: problem, treatment, test, occurrence, and evidential).

- Boundary adjustment post-processing is significantly helpful for exact boundary identification. (Events considered: problem, treatment, and test).

### A. Future work

Potential extensions of work presented in this paper may explore:

- Boundary adjustment post-processing applied on IO and IOB models.

- Boundary adjustment by use of semantic resources during post-processing.

- Explore conditional probability thresholds .

## VII. APPENDIX

Table I and Table II provide statistics from the union of the training and test datasets of the i2b2 2012 event corpus:

TABLE I.  EVENT DISTRIBUTION

| Event | Instances | Proportion |
|---|---|---|
| Problem | 9331 | 32.93 % |
| Test | 4769 | 16.83 % |
| Treatment | 7114 | 25.11 % |
| Occurrence | 5784 | 20.42 % |
| Evidential | 1334 | 4.71 % |
| **Total** | 28332 | 100 % |

TABLE II. ACRONYM/ABBREVIATION DISTRIBUTION

| Event | % Acronyms/Abbreviations | Proportion |
|---|---|---|
| Problem | 9 | 55 % |
| Test | 24 | 37 % |
| Treatment | 11 | 43 % |
| Occurrence | 1 | 30 % |
| Evidential | 0 | 10 % |

### ACKNOWLEDGMENT

I would like to thank John Keane for his guidance.

The University of Manchester PhD Studentship, UK Engineering and Physical Science Research Council (ESPRC), Christie Paediatric Oncology Charitable Fund.